# SPATIO-TEMPORAL GRAPHICAL MODEL SELECTION[*]

By Patrick Harrington[†], Alfred Hero[†]

*University of Michigan*[†]

We consider the problem of estimating the topology of spatial interactions in a discrete state, discrete time spatio-temporal graphical model where the interactions affect the temporal evolution of each agent in a network. Among other models, the susceptible, infected, recovered ($SIR$) model for interaction events fall into this framework. We pose the problem as a structure learning problem and solve it using an $\ell_1$-penalized likelihood convex program. We evaluate the solution on a simulated spread of infectious over a complex network. Our topology estimates outperform those of a standard spatial Markov random field graphical model selection using $\ell_1$-regularized logistic regression.

**1. Introduction.** This paper treats the problem of learning the interaction structure of a spatio-temporal graphical model for a discrete state and discrete time stochastic process known as the susceptible, infected, recovered ($SIR$) model. The presence of spatial interactions cause adjacent nodes in the graph to affect each others states over time. Learning the topology of this graph is known as model selection. We cast this graphical model selection problem as a penalized likelihood problem, resulting in a convex program for which convex optimization solvers can be applied. $SIR$ spatio-temporal graphical models are commonly used in modeling the random propagation of information between nodes in large networks in bioinformatics, signal processing, public health, and national security ([4]; [9]; [21]). Knowing the network link structure allows accurate prediction of individual node states and can aid the development of control and intervention strategies for such networks. This paper develops a tractable method to estimate the topology of the network for the $SIR$ spatio-temporal graphical model from empirical data.

Exact solutions of the graphical model selection problem is NP hard due to the combinatorial nature of enumeration through the discrete space of possible graph topologies. Researchers studying Bayesian networks, both

[*]This research was partially supported by the AFRL ATR Center through a Signal Innovations Group subcontract, grant number SIG FA8650-07-D-1221-TO1

static and dynamic, have developed exact and approximate methods for selecting a good candidate topology (5; 8; 20). Such methods are appropriate for networks of small size and of unknown generative models for the observations. However, they are difficult to scale to larger graphs. $SIR$ processes are used to model transmission events on complex networks which tend to be sparse in their interactions (22; 3; 21; 7; 4), so that there are relatively few edges in the graph. Over the past decade sparse regularization methods have been developed for graphical model selection using $\ell_1$-regularization and other approaches. Examples include Gaussian graphical models (GMMs) (18; 10; 33; 26; 24) and Markov random fields (MRFs) (16; 31; 26).

The $SIR$ model used throughout this paper is both discrete state and discrete time and thus any $\ell_1$-penalized GMM method that is designed for real valued Gaussian random vectors would not be appropriate for this model. The structure learning algorithms for MRFs discussed in (16; 31; 26) are designed for discrete samples drawn from a MRF and most are limited to binary states (the $SIR$ model has three states and is a different generative model than the MRF). Research in MRFs and GMMs have successfully used the $\ell_1$-penalty to control the sparseness of the estimated graphical model topologies and we will adopt this approach for the $SIR$ model. The method presented in this paper also scales to large networks more easily than traditional Bayesian network structure learning algorithms (5; 8; 20).

The proposed sparse structure learning method is designed for graphs that incorporate random causal transmission events affecting the evolution of the node states, such occurs in the propagation of infectious disease. Identifying the structure of social networks in tracking epidemics has received increased attention due to the global response to pandemic influenza A (H1N1) 2009. We illustrate the accuracy of the proposed network structure learning on two moderate sized complex networks using real-world epidemiological parameters that approximate an H1N1 flu inspired outbreak (32). We compare performance of the proposed estimation method against a MRF graphical model selection using $\ell_1$-regularized logistic regression (31). The proposed method is more accurate than generic approaches such as (31) for detection of anomalous network structure given sampled data from a spatio-temporal $SIR$ process.

**2. SIR Spatio-Temporal Graphical Models.** The $SIR$ graphical model has been used to approximate the general problem of modeling the evolution of node states in a network in which there is random transmission of disease or information between adjacent nodes on a graph (3; 21; 7; 4). In the limit of large populations with equal mixing rates, $SIR$ models have been used to model population proportions of particular states using differential equations (1; 25; 19; 13). Unlike these studies, this paper addresses the problem of estimation of the topology of interactions between individual nodes in the network.

The $SIR$ graphical model is a discrete time, discrete state model for the states of nodes in the network. Nodes can only affect the states of adjacent nodes in the network when they are in the "infected" state. The state of a node is given by $X_{i,k}$, where $i$ refers to the individual (node) and $k$ denotes time, and $X_{i,k}$ takes on values $x \in \{0, 1, 2\}$ (corresponding to "susceptible", "infected", and "recovered", respectively). The model is specified by the state transition probabilities given in the 3x3 stochastic matrix

$$(2.1) \qquad P_{i,k|k-1} = \begin{bmatrix} 1 - q_{i,k|k-1} & 0 & \gamma \\ q_{i,k|k-1} & 1 - \alpha & 0 \\ 0 & \alpha & 1 - \gamma \end{bmatrix}$$

where $q_{i,k|k-1}$ is the probability of transmission from "infected" neighbors of node $i$ at time $k$, $\gamma$ is the probability that node $i$ transitions from "recovered" to "susceptible", and $\alpha$ is the probability that node $i$ transitions from "infected" to "recovered". Since (2.1) allows a transition from recovered back to susceptible, this is actually a $SIRS$ model ($SIR$ and $SIRS$ will be used interchangeably to refer to the three state stochastic process). For $p$ nodes, the spatial topology of the network is defined by the interconnectivity, or adjacency, matrix

$$(2.2) \qquad \mathcal{E} = \begin{bmatrix} \mathcal{E}_{1,1} & \cdots & \mathcal{E}_{1,p} \\ \vdots & \ddots & \vdots \\ \mathcal{E}_{p,1} & \cdots & \mathcal{E}_{p,p} \end{bmatrix}$$

where the $l, m^{th}$ entry $\mathcal{E}_{l,m} \in \{0, 1\}$ is the indicator event that nodes $l$ and $m$ are connected. The pattern of non-zero entries in (2.2) specifies the interconnection topology of the network. The fundamental assumptions for an $SIR$ network model is that the transition probabilities do not depend on node $i$ while the interconnectivity matrix (2.2) is independent of time $k$. Under these assumptions, the joint distribution of an observed trajectory of length $T$, represented by the $p$-dimensional discrete state vector $X_k = [X_{1,k}, \ldots, X_{p,k}]^T$, factorizes

$$(2.3)$$
$$\mathbb{P}(X_1, \ldots, X_T) = \prod_{k=2}^{T} \mathbb{P}(X_k | X_{k-1}) = \prod_{k=2}^{T} \prod_{i=1}^{p} \mathbb{P}\left(X_{i,k} | \{X_{j,k-1}\}_{j \in \{\eta(i), i\}}\right),$$

where the neighborhood of node $i$ is denoted

$$(2.4) \qquad \eta_i = \{j : \mathcal{E}_{i,j} \neq 0\}.$$

The core component of most variations of the $SIR$ model is the assumption that node $i$ is conditionally independent of all non-neighboring nodes given the states of node $i$ and its neighbors at time $k-1$. Each neighbor can transmit the "infection" to node $i$ independent of the others neighbors. Under these assumptions, the probability of at least one transmission to node $i$ at time $k$ is given by

$$(2.5) \qquad q_{i,k|k-1} = 1 - \prod_{j \in \eta_i} \left(1 - \omega z_{j,k-1}^{(1)}\right),$$

where $z_{k-1}^{(1)} \in \{0, 1\}$ is the indicator random variable of the $j^{th}$ variable being in state "infected" at previous time $k-1$ and $\omega$ is the prior Bernoulli probability of transmission between $j$ and $i$ (also referred to as the attack rate). The conditional transition distribution in (2.3) is given by the following multinomial distribution

$$(2.6) \qquad \mathbb{P}\left(X_{i,k} = x | \{X_{j,k-1} = x_{j,k-1}\}_{j \in \{\eta_i, i\}}\right) = \prod_{x \in \{0,1,2\}} \left(p_{i,k|k-1}(x)\right)^{z_{i,k}^{(x)}}$$

with indicator variable $z_{i,k}^{(x)} = \mathbb{I}_{\{x_{i,k}=x\}}$ and label probability $p_{i,k|k-1}(x)$ given by

$$p_{i,k|k-1}(x) = \begin{cases} \gamma z_{i,k-1}^{(2)} + z_{i,k-1}^{(0)} \prod_{j \in \eta_i} \left(1 - \omega z_{j,k-1}^{(1)}\right) & , x = 0 \\ z_{i,k-1}^{(0)} \left(1 - \prod_{j \in \eta_i} \left(1 - \omega z_{j,k-1}^{(1)}\right)\right) + (1-\alpha) z_{i,k-1}^{(1)} & , x = 1 \\ \alpha z_{i,k-1}^{(1)} + (1-\gamma) z_{i,k-1}^{(2)} & , x = 2, \end{cases}$$

where the model parameters are defined in (2.1). While the proposed graphical model selection method in this paper is motivated using the canonical three state $SIR$ model, the method can be extended to any discrete state, discrete time stochastic model with state interactions of the form of the probability of transmission given in (2.5).

## 3. Spatio-Temporal Topology Estimation.
Here we develop an estimate of the topology $\mathcal{E}$ (2.2) given training sequences $\mathcal{D}$ of observed states

$$(3.1) \qquad \mathcal{D} = \{x_{i,k}\}_{i=1,k=1}^{p,T},$$

where $T$ is the horizon of the measurement period.

It will be convenient to rewrite the term involving the probability of transmission in (2.5) as

$$
\begin{aligned}
\prod_{j \in \eta(i)}\left(1-\omega z_{j, k-1}^{(1)}\right) & =\exp \left\{\log \left(\prod_{j \in \eta_i}\left(1-\omega z_{j, k-1}^{(1)}\right)\right)\right\} \\
& =\exp \left\{\sum_{j \in \eta_i} \log \left(1-\omega z_{j, k-1}^{(1)}\right)\right\} \\
& =\exp \left\{\sum_{j \in \eta_i} \log (1-\omega) z_{j, k-1}^{(1)}\right\},
\end{aligned}
$$

where we have exploited the fact that $\log (1-\omega z_{j, k-1}^{(1)})=\log (1-\omega) z_{j, k-1}^{(1)} \leq 0$ in (3.2). Define $\theta_{i, j}$

$$
\theta_{i, j}= \begin{cases}\log (1-\omega) & , \mathcal{E}_{i, j}=1 \\ 0 & , \mathcal{E}_{i, j}=0\end{cases}
$$

and re-writing the sum term in (3.2) to run over the other $p-1$ nodes we arrive at the following

$$
(3.3) \quad 1-q_{i, k \mid k-1}=\exp \left\{\sum_{j \neq i} \theta_{i, j} z_{j, k-1}^{(1)}\right\}, \; \theta_{i, j} \in\{\log (1-\omega), 0\} \; \forall j \neq i .
$$

Inserting (3.3) into the state label probabilities, we have

$$
p_{i, k \mid k-1}(x)= \begin{cases}\gamma z_{i, k-1}^{(2)}+z_{i, k-1}^{(0)} e^{\sum_{j \neq i} \theta_{i, j} z_{j, k-1}^{(1)}} & , x=0 \\ z_{i, k-1}^{(0)}\left(1-e^{\sum_{j \neq i} \theta_{i, j} z_{j, k-1}^{(1)}}\right)+(1-\alpha) z_{i, k-1}^{(1)} & , x=1 \\ \alpha z_{i, k-1}^{(1)}+(1-\gamma) z_{i, k-1}^{(2)} & , x=2 .\end{cases}
$$

Define the $p-1$ dimensional column vector $\theta_i$ by $\theta_i=\{\theta_{i, j}\}_{j \neq i}$. Given the spatial and temporal conditional independence assumptions represented in (2.3), the joint likelihood can be written as the multinomial distribution

$$
(3.4) \quad \mathcal{L}(\phi ; \mathcal{D})=\prod_{k=2}^T \prod_{i=1}^p \prod_{x \in\{0,1,2\}}\left(p_{i, k \mid k-1}(x)\right)^{z_{i, k}^{(x)}}
$$

with $\phi = \{\theta, \alpha, \gamma, \omega\}$ and $\theta = \{\theta_i\}_{i=1}^p$. The joint log-likelihood can be written as

$$\ell(\phi; \mathcal{D}) = \sum_{i=1}^p \ell(\phi_i; \mathcal{D}), \tag{3.5}$$

with $\phi_i = \{\theta_i, \alpha, \gamma, \omega\}$. The objective is to estimate the topology parameter $\theta$ while the $\alpha$, $\gamma$, and $\omega$ are nuisance parameters. The $i^{th}$ log-likelihood function is

$$
\begin{aligned}
\ell(\theta_i; \mathcal{D}) &= \sum_{k=2}^T \left\{ z_{i,k}^{(0)} \log p_{i,k|k-1}(0) + z_{i,k}^{(1)} \log(p_{i,k|k-1}(1)) \right\} \\
&= \sum_{k=2}^T \left\{ z_{i,k|k-1}^{(0,0)} \sum_{j \neq i} \theta_{i,j} z_{j,k-1}^{(1)} + z_{i,k|k-1}^{(1,0)} \log\left(1 - e^{\sum_{j \neq i} \theta_{i,j} z_{j,k-1}^{(1)}}\right) \right\},
\end{aligned}
\tag{3.6}
$$

with $z_{i,k|k-1}^{(0,0)} = z_{i,k}^{(0)} z_{i,k-1}^{(0)}$ and $z_{i,k|k-1}^{(1,0)} = z_{i,k}^{(1)} z_{i,k-1}^{(0)}$. Note that (3.6) only includes the state transition probabilities that involve $\theta_{i,j}$ since $\theta_{i,j}$ is obtained by optimizing over $\ell(\theta_i; \mathcal{D})$. In particular, the transition from any state to recovered does not depend on $\theta_{i,j}$. Note that the only parameter appearing in (3.6) necessary for estimation of $\theta$ is the transmission attack rate $\omega$, appearing implicitly through the definition of $\theta_{i,j}$, $\theta_{i,j} \in \{\log(1-\omega), 0\}$.

Maximization of the likelihood over all possible $\theta \in \{\log(1-\omega), 0\}^{p(p-1)}$ is intractable even for small networks. The key to our maximum likelihood estimation approach is to relax $\theta_{i,j}$ to a continuous valued variable lying between its discrete values $\log(1-\omega)$ and $0$, i.e., $\log(1-\omega) \leq \theta_{i,j} \leq 0$.

We use an $\ell_1$-penalty on the likelihood to enforce sparsity, i.e., only a few $\theta_{i,j}$ are non-zero. Such $\ell_1$-penalization is common in high dimensional statistical problems (28; 31; 14; 18; 10; 33; 26). This yields the following convex program

$$
\begin{aligned}
&\min_\theta -\ell(\theta; \mathcal{D}) + \lambda \|\theta\|_{\ell_1} \\
&\text{s.t. } \log(1-\omega) \preceq \theta \preceq 0
\end{aligned}
\tag{3.7}
$$

with $\lambda > 0$ and $\preceq$ denotes element wise inequality between vectors. The estimated neighborhood set of node $i$ is then

$$\hat{\eta}_i(\lambda) = \{j : \hat{\theta}_{i,j}(\lambda) < 0\}. \tag{3.8}$$

The set of all such neighborhoods will specify a (directed) graph that can be used to estimate the network topology $\mathcal{E}$ in (2.2). Specifically, the estimate of the $l^{th}$ $m^{th}$ entry of $\mathcal{E}$ by $\hat{\mathcal{E}}_{l,m}(\lambda) = \mathbb{I}_{\{\hat{\theta}_{l,m}(\lambda) < 0\}}$. The global estimate of the topology is then defined as $\hat{\mathcal{E}}(\lambda) = \{\hat{\mathcal{E}}_{l,m}(\lambda)\}_{l,m}$.

3.1. *Incorporating Prior Knowledge.* There generally exists prior topological constraints that couple the optimization over $\{\theta_i\}_{i=1}^p$ for different $i$ in (3.5). One such topological constraint is symmetry in the interactions, i.e., $\theta_{i,j} = \theta_{j,i}$, corresponding to an undirected graph $\mathcal{E}$. One way to incorporate this symmetry is to use augmented lagrangian methods that impose symmetry in the form of a variational penalty, e.g., $\sum_{i,j}(\theta_{i,j} - \theta_{j,i})^2$ (23). Another method is to relax the symmetry constraint during the optimization followed by averaging the $\theta_{i,j}$ and $\theta_{i,j}$ together after optimization is completed.

If symmetry in $\theta_{i,j}$ is not imposed, the joint log-likelihood naturally factorizes as in (3.5), and can be decoupled by applying a coordinate descent-like likelihood function maximization that cycles through different nodes, updating its neighborhoods and holding the other $\theta_i$'s fixed:

$$\min_{\theta_i} -\ell(\theta_i; \mathcal{D}) + \lambda \sum_{j \neq i} |\theta_{i,j}|$$

(3.9)          subject to $\log(1 - \omega) \leq \theta_{i,j} \leq 0, \ \forall j \neq i.$

Researchers may have additional prior knowledge such as known interactions, known non-interactions, or minimum or maximum size of neighborhoods. Some common forms of prior knowledge, and their corresponding constraints are summarized in Table 1.

TABLE 1
*Common prior knowledge for complex networks appearing as constraints for the SIR graphical model selection problem*

| Prior Knowledge | Form of Constraint |
|---|---|
| Symmetry | $\theta_{i,j} = \theta_{j,i}$ |
| Known Interactions | $\theta_{i,j} = \log(1 - \omega), j \in \eta_i$ |
| Known Non-Interactions | $\theta_{i,j} = 0, j \notin \eta_i$ |
| Min Possible Size of Neighborhood | $\sum_{j \neq i} \theta_{i,j} \geq a_i \cdot \log(1 - \omega)$ |
| Max Possible Size of Neighborhood | $\sum_{j \neq i} \theta_{i,j} \leq b_i \cdot \log(1 - \omega)$ |

It is more natural to work with the dual of the objective function in (3.7). In the dual one can immediately identify which of the inequality constraints are active. For instance, if one has prior knowledge regarding the maximum size of a particular neighborhood, e.g., $\sum_{j \neq i} \theta_{i,j} \leq b \cdot \log(1 - \omega)$, one can determine if $b \cdot \log(1 - \omega) < s$, in which case, the constraint of $\|\theta\|_{\ell_1} \leq s$ would be inactive for the subvector $\theta_i$. This results in convexity preserving topological constraints

$$
(3.10) \qquad \min_{\theta} - \ell(\theta; \mathcal{D})
$$
$$
\text{subject to} \quad \|\theta\|_{\ell_1} \leq s
$$
$$
\log(1 - \omega) \preceq \theta \preceq 0
$$
$$
\{h_j(\theta) \leq \nu_j\}_{j=1}^{k}
$$
$$
\{g_l(\theta) = 0\}_{l=1}^{r}.
$$

3.2. *Numerical Solution.* The proposed $\ell_1$-penalized likelihood problem in (3.9) is a convex program where there exists a variety of powerful solvers capable of producing a solution (2). The proposed numerical solution in this paper is most appropriate for networks on the order of hundreds to a few thousand nodes. For networks on the order of tens of thousands of nodes, a large scale method such as the one given in (14) might be more appropriate.

We will relax the symmetry constraints when optimizing over $\theta$ and later impose them as a post-estimation heuristic

$$
(3.11) \qquad \hat{\eta}_i(\lambda^*) \leftarrow \hat{\eta}_i(\lambda^*) \cup j, \text{if } i \in \hat{\eta}_j(\lambda^*) \cap j \notin \hat{\eta}_i(\lambda^*) \forall i, j.
$$

We use a coordinate-wise gradient descent based method for solving (3.9) by quadratically expanding the negative log-likelihood, resulting in iteratively solving a sequence of quadratic programs that incorporates an additional line search. The Newton-step update is accomplished by solving

$$
\delta\theta_i^{(m)} = \arg\min_{\theta_i} \frac{1}{2}\theta_i^T H_i^{(m)} \theta_i + \theta_i^T g_i^{(m)} + \lambda \sum_{j \neq i} |\theta_{i,j}|
$$
$$
(3.12) \qquad \text{s.t.} \quad \log(1 - \omega) \leq \theta_{i,j} \leq 0, \ \forall j \neq i,
$$

with gradient

$$
(3.13) \qquad g_i^{(m)} = -\nabla \ell(\theta_i; \mathcal{D})\big|_{\theta_i = \hat{\theta}_i^{(m)}},
$$

and Hessian

$$
(3.14) \qquad H_i^{(m)} = -\nabla^2 \ell(\theta_i; \mathcal{D})\big|_{\theta_i = \hat{\theta}_i^{(m)}}.
$$

The updated parameter $\hat{\theta}_i^{(m+1)}$ given by

$$(3.15) \qquad \hat{\theta}_i^{(m+1)} = \hat{\theta}_i^{(m)} + \epsilon_i^{(m)} \delta\theta_i^{(m)},$$

with step size $\epsilon_i^{(m)}$ determined by performing a backtracking line search (2)

$$(3.16)$$
$$\text{while } -\ell(\hat{\theta}_i^{(m)} + \epsilon_i^{(m)} \delta\theta_i^{(m)}; \mathcal{D}) > -\ell(\hat{\theta}_i^{(m)}; \mathcal{D}) + 0.2\epsilon_i^{(m)}(g_i^{(m)})^T \delta\theta_i^{(m)}, \epsilon_i^{(m)} \leftarrow 0.3\epsilon_i^{(m)},$$

with $\epsilon_i^{(m)}$ initially set to 1. While (3.12) is convex, the presence of the $\ell_1$-norm makes the objective function non-differentiable. However, the objective function can be transformed into an equivalent convex, differentiable objective by replacing the $\ell_1$-norm with linear inequality constraints (2; 14). An alternative to solving the Newton update (3.12) with the $(p-1)\text{x}(p-1)$ Hessian is replace it with a quasi Newton update which construct a surrogate objective function (29; 17; 15) and replaces the Hessian, $H_i^{(m)}$ with $\alpha_i^{(m)} I$, where $I$ is the identity and $\alpha_i^{(m)}$ is chosen such that

$$(3.17) \qquad \alpha_i^{(m)} I \succeq H_i^{(m)},$$

and (3.17) means that $\alpha_i^{(m)} I - H_i^{(m)} \succeq 0$ is positive semi-definite. A consequence of the proposed penalized likelihood formulation for the $SIR$ model is that $H_i^{(m)}$, in addition to being symmetric and positive semi-definite, has positive entries, i.e., $(H_i^{(m)})_{s,r} \geq 0$. Such non-negative conditions on the entries in $H_i^{(m)}$ can be enforced by using the Perron-Frobenius bound (12)

$$(3.18) \qquad \max_s \lambda_s \left( H_i^{(m)} \right) \leq \max_s \sum_r \left( H_i^{(m)} \right)_{s,r},$$

where the optimization is applied to the upper bound

$$(3.19) \qquad \alpha_i^{(m)} = \max_s \sum_r \left( H_i^{(m)} \right)_{s,r},$$

thus guaranteeing (3.17).

By replacing the Hessian with a diagonal surrogate is that the $p-1$-dimensional quadratic program in (3.12) factorizes into $p-1$ individual programs which have an analytical update and can be evaluated simultaneously. The update for $\theta_{i,j}$ under such a surrogate Hessian becomes

$$\delta\theta_{i,j}^{(m)} = \arg\min_{\theta_{i,j}} \frac{1}{2}\alpha_i^{(m)}\theta_{i,j}^2 + g_{i,j}^{(m)}\theta_{i,j} + \lambda|\theta_{i,j}|, \ \log(1-\omega) \leq \theta_{i,j} \leq 0$$

$$(3.20) \qquad = \begin{cases} \frac{-1}{\alpha_i^{(m)}} \left( |g_{i,j}^{(m)}| - \lambda \right)_+ & : g_{i,j}^{(m)} < \lambda - \alpha_i^{(m)} \cdot \log(1-\omega) \\ \log(1-\omega) & : g_{i,j}^{(m)} \geq \lambda - \alpha_i^{(m)} \cdot \log(1-\omega) \end{cases}$$

with $(u)_+ = \max(0, u)$. The proposed gradient descent method for the $\ell_1$-penalized likelihood problem for the spatio-graphical model selection problem is summarized in Algorithm 1 (below).

## Algorithm 1

1. Let $\hat\theta_i^{(0)} \in [\log(1-\omega), 0]^{p-1}$ be an initial parameter vector for the $i^{th}$ neighborhood.
2. Update $\delta\theta_i^{(m)}$ by solving (3.12) or solving (3.20) $\forall j \neq i$ with surrogate diagonal Hessian given by (3.19)
3. $\epsilon_i^{(m)} \leftarrow$ backtracking line search from (3.16)
4. $\hat\theta_i^{(m+1)} = \hat\theta_i^{(m)} + \epsilon_i^{(m)} \delta\theta_i^{(m)}$
5. If convergence criteria met, stop and repeat step 1 with next node index, $i \leftarrow i + 1$. If convergence criteria not met, update gradient and Hessian (and potentially the surrogate diagonal Hessian) and repeat step 2 through 5 Note: Algorithm 1 can be parallelized across all $p$ log-likelihoods rather than the cyclical update of $i \leftarrow i + 1$. Symmetry is imposed through (3.11).

A possible speed up would be to perform active set updates to those coefficients which are non-zero by preferentially updating the coefficients corresponding to nodes that most likely belong to the neighborhood. Such active set updates have been used successfully in estimating sparse partial correlations (24). They have also been proposed to block co-ordinate descent in group lasso logistic regression (17). Implementing such accelerations is out of the scope of this paper.

3.3. *Selection of Tuning Parameters.* Algorithm 1 requires specification of the tuning parameter $\lambda$. Typically, an estimate of the best $\lambda$ is desirable in order to perform cross validation or other error assessment. In this paper we report a BIC-like penalty, similarly used in previous work on the estimation of partial correlation networks (24), for selecting the best estimate of $\lambda$, denoted by $\lambda^*$, by cross validation. Specifically, assuming the attack rate $\omega$ is known, we perform the update $\theta_{i,j}$ as follows

$$(3.21) \qquad \hat\theta_{i,j}(\lambda) \leftarrow \log(1-\omega), \forall i,j \in \{i,j : \hat\theta_{i,j}(\lambda) < 0\}.$$

The BIC penalty for the $i^{th}$ node is

$$(3.22) \qquad BIC_i(\lambda) = -\ell_i(\hat\theta_i(\lambda); \mathcal{D}) + \frac{1}{2}\log T_i \ \#\{j : \hat\theta_{i,j}(\lambda) < 0\},$$

where $\#\{i,j : \hat\theta_{i,j}(\lambda) < 0\}$ is the number of non-zero entries in the estimator. The term $T_i = \#\{k : z_{i,k}^{(0)} = 1\}$ represents the effective time horizon for the $i^{th}$ node as the number of terms in the $i^{th}$ log-likelihood, which depends on the number of of $z_{i,k}^{(0)}$ equal to one (see (3.6)). Given (3.22), there will be multiple regularization parameters, one for each neighborhood $i$:

$$(3.23) \qquad \lambda_i^* = \arg\min_\lambda BIC_i(\lambda).$$

The common approach is to impose that all the $\lambda_i^*$'s are the same and solve for a single tuning parameter

$$(3.24) \qquad \lambda^* = \arg\min_\lambda \sum_{i=1}^p BIC_i(\lambda).$$

The latter approach has been previously used in controlling the sparseness of estimated partial correlation networks (24) and learning directed acyclic graphs (DAGs) (27).

**4. Numerical Results.** Given the global response to the recent outbreak of pandemic influenza A (H1N1) 2009, the ability of public health organizations and world governments to develop effective control and intervention strategies depends on knowledge of the topology of social networks. We illustrate the proposed penalized likelihood topology estimate for the problem of identifying the structure of synthetic social networks given disease spread that has attack rate parameters that simulate H1N1, specifically $\omega = 0.273$ as reported in (32). The other two parameters, not needed for network inference but necessary for generating $SIR$ trajectories from (2), were taken as $\alpha = 0.250$ reflecting a mean infectious period of 4 days and $\gamma = 0.100$ producing an average time of 10 days for transition from "recovered" to "susceptible".

We simulated two 200 node networks using two types of connection models: scale-free and small-world. These models have been proposed for many practical complex networks (22). The two randomly generated networks used for experiments were created using the iGraph package for $R$ (6). The power law network was sampled such that the degree distribution reflected those which appear in real complex networks. Specifically, the exponent parameter of the degree distribution was taken as 2.2, consistent with evidence reported in (22). The rewiring probability of the small-world network was taken as 0.1 to elicit tight communities that were loosely connected to other clusters.

The $SIR$ model (2) was used to generate training, validation, and test data for each of the two simulated 200 node networks. The networks were initialized with 40 randomly selected nodes were in "infected" state while the rest were in "susceptible" state. The quadratic program appearing in Algorithm 1, (3.12), was solved using the CVX environment in MATLAB and the solver SDPT3 4.0 (11; 30) with cold start initializations of $\hat{\theta}_i^{(0)} = 0$. Symmetry was included in the estimated neighborhoods following the post estimation heuristic (3.11).

We present a comparison against a modified version of graphical model selection using $\ell_1$-logistic regression ($\ell_1$-LR) (31). Since the method described in (31) is designed for binary random variables generated from an Ising model, to implement $\ell_1$-LR we transform the three state $SIR$ variables to binary random variables. The transformation is the following: for each node $i$, the indicator event of the $i^{th}$ node transitioning from "susceptible" to "infected", is regressed on all other $p-1$ "infected" nodes indicator variables at previous time $k-1$ with a bias controlling constant as explained in (31) and symmetry imposed through (3.11). By transforming the multi-state $SIR$ random variables to the binary random variables for the implementation of $\ell_1$-LR, we capture the causality of transmission from neighbors. While we transform the three state $SIR$ random variables to two state random variables for implementing $\ell_1$-LR, the proposed graphical model selection in this paper, referred to as $\ell_1$-SIR, uses the original three state variables in the log-likelihood (3.6). As the estimated parameters using $\ell_1$-LR (31) can take on any value on the real-line, we define the estimated neighborhood for the $i^{th}$ node as those estimates with non-zero value.

The ROC curves corresponding to $\ell_1$-SIR and the modified $\ell_1$-LR for the scale-free network and small-world network for $T = \{500, 1000\}$ are displayed in Figure 1(a) and Figure 1(b), respectively. Inspection of Figure 1 validates that the proposed $\ell_1$-SIR graphical model selection outperforms $\ell_1$-LR when confronted with data drawn from the $SIR$ distribution. At a false alarm rate of 5%, we see that the proposed $\ell_1$-SIR method achieves a $5\% - 10\%$ gain in power over the modified $\ell_1$-LR method for both networks. As $\ell_1$-LR (31) uses one $\ell_1$ penalty, for baseline comparison between these two graphical model selection algorithms, only a single regularization penalty was used in the ROC curves generated from $\ell_1$-SIR. Both structure learning methods perform poorer in the case of the small-world network than in the case of the scale-free network. This is possibly due to the increased frequency of re-infection in the tight clusters of the small-world network.

We next present the model selection performance on the 200 node scale-free network using the proposed method with global and neighborhood specific penalties, optimized by minimizing the BIC penalties (3.24) and (3.23), respectively, for time durations of $T = \{100, 400, 700, 1000\}$. The images in Figures 2 and 3 reflect the estimated network topologies, represented as symmetric adjacency matrices $\mathcal{E}$, averaged over the 1000 resampled initial conditions corresponding $T = 100$ and $T = 1000$, respectively. Subfigures a.) through c.) correspond to ground truth, $\ell_1$-SIR with a single $\ell_1$-penalty, and $\ell_1$-SIR with neighborhood specific $\ell_1$-penalty, respectively.

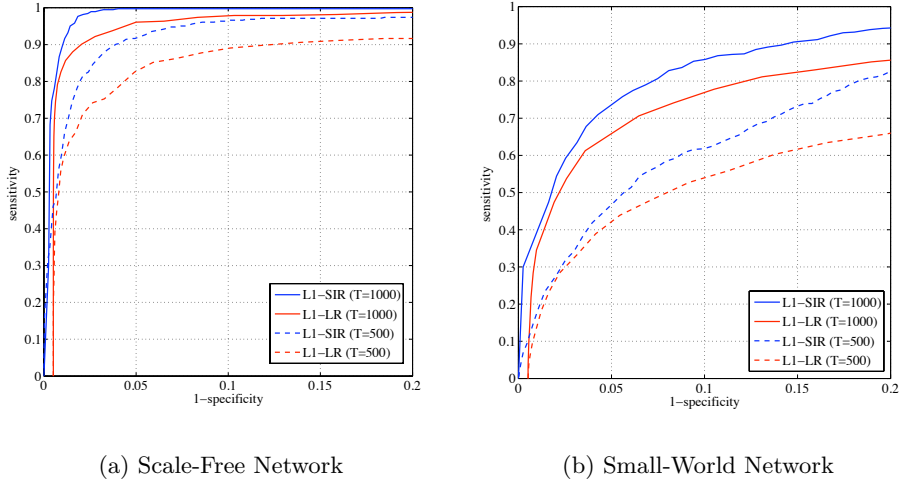

(a) Scale-Free Network          (b) Small-World Network

Fig 1. *ROC curves of $\ell_1$-SIR graphical model selection* **(blue)** *vs. $\ell_1$-logistic regression* **(red)** *for number of time points $T = \{500, 1000\}$*

The intensity, located at row $i$ and column $j$, indicates the frequency of an edge discovered between nodes $i$ and $j$, white designates a strong edge and black designates no edge. Visual inspection of these figures establish that the proposed $\ell_1$-SIR graphical model selection methods accurately extract the global community structure of the scale-free network when using a single or multiple penalties to enforce sparseness.

A quantitative comparison of accuracy of topology estimation is given by the sensitivity, specificity and probability of error. Table 2 summarizes the mean (with standard deviation shown in parentheses) when assessing the performance across the 1000 reconstructed topologies corresponding to the 1000 resampled simulations. We see that the sensitivity of this method, using a single $\lambda^*$ and multiple $\{\lambda_i^*\}_{i=1}^p$, increases when the number of time samples increases while the specificity remains robust to the number of time samples and consistently above 0.96. Likewise, the global probability of error is below 0.05 for both methods across all time horizons explored. It is worth noting that the proposed method is only able to resolve an interaction between nodes $i$ and $j$ if both nodes states have changed at some point throughout

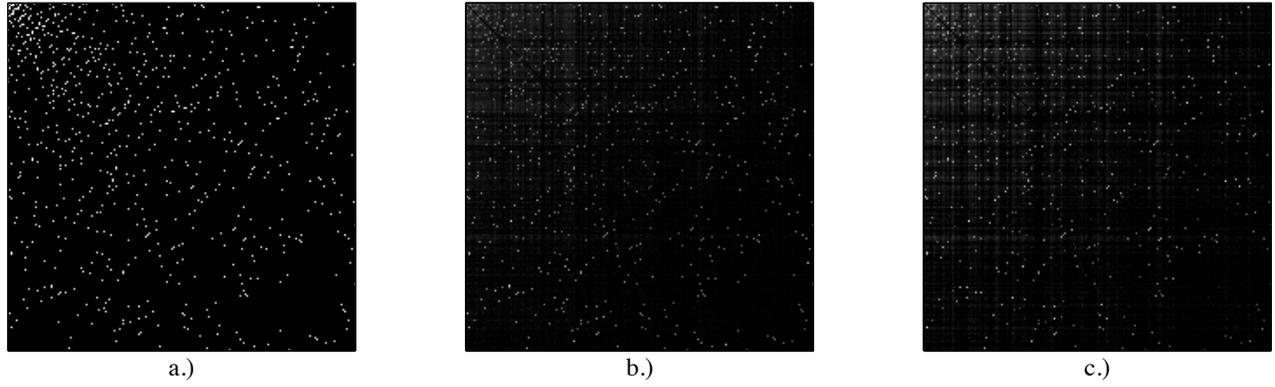

Fig 2. *% zeros in the reconstruction of edges in 200 node synthetic scale free network under 100 time points resampled over 1000 initial conditions of 40 randomly selected nodes as "infected" with rest "susceptible". a.) ground truth, b.) single tuning parameter, c.) multiple tuning parameters (white - 0% black 100%)*

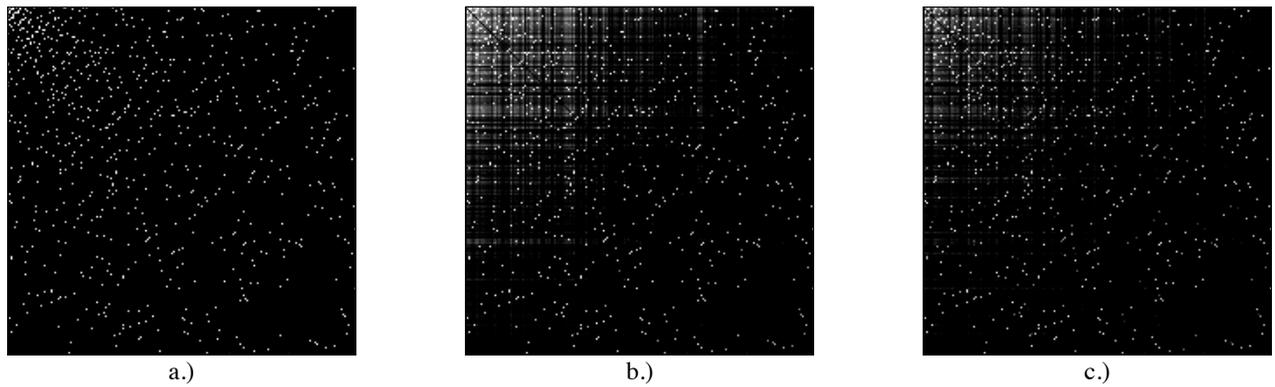

Fig 3. *% zeros in the reconstruction of edges in 200 node synthetic scale free network under 1000 time points resampled over 1000 initial conditions of 40 randomly selected nodes as "infected" with rest "susceptible". a.) ground truth, b.) single tuning parameter, c.) multiple tuning parameters (white - 0% black 100%)*

the monitoring interval. Therefore for small time horizons, the epidemic may not have enough time to propagate the entire graph thus inhibiting the ability to accurately detect interactions.

A scale-free network has a wide distribution of vertex degrees (few hubs, many lesser connected nodes). Figure 4 a.), b.), and c.) show the sensitivity, specificity, and probability of error, respectively, of correctly detecting the neighborhood of each node as a function of increasing vertex degree. In all three subfigures, we see that regularizing with tuning parameters characteristic to each neighborhood $\{\lambda_i^*\}_{i=1}^p$ selected according to (3.23) tends to produce similar sensitivity and specificity with lower probability of error across all types of node degrees than when regularizing with a single penalty $\lambda^*$ selected according to (3.24).

TABLE 2

*Detection statistics vs. time horizon for 200 node synthetic scale-free network with trajectories resampled over 1000 initial conditions of 40 randomly selected nodes as "infected" with rest "susceptible"*

| Method | $T$ | $Sens.(\lambda^*)$ | $Spec.(\lambda^*)$ | $P_e(\lambda^*)$ |
|---|---|---|---|---|
| $\ell_1$-SIR($\lambda^*$) | 100 | 0.40(0.02) | 0.96(0.00) | 0.05(0.00) |
| $\ell_1$-SIR($\{\lambda_i^*\}_{i=1}^p$) | 100 | 0.34(0.02) | 0.97(0.00) | 0.05(0.00) |
| $\ell_1$-SIR($\lambda^*$) | 400 | 0.80(0.05) | 0.97(0.00) | 0.03(0.00) |
| $\ell_1$-SIR($\{\lambda_i^*\}_{i=1}^p$) | 400 | 0.78(0.05) | 0.96(0.00) | 0.04(0.00) |
| $\ell_1$-SIR($\lambda^*$) | 700 | 0.95(0.08) | 0.96(0.00) | 0.04(0.00) |
| $\ell_1$-SIR($\{\lambda_i^*\}_{i=1}^p$) | 700 | 0.95(0.07) | 0.96(0.00) | 0.04(0.00) |
| $\ell_1$-SIR($\lambda^*$) | 1000 | 0.97(0.08) | 0.96(0.00) | 0.03(0.00) |
| $\ell_1$-SIR($\{\lambda_i^*\}_{i=1}^p$) | 1000 | 0.97(0.08) | 0.96(0.00) | 0.03(0.00) |

The performance of the proposed method was also assessed for a 200 node small-world network. Visual inspection of Figure 6 shows that the proposed method method accurately extracts the small-world community structure, represented by the recovery of the banded structure of the adjacency matrices. In addition to detecting the characteristic clusters of the small-world ground truth network, the method also tends to identify the between-cluster interactions which are depicted in the off-diagonal elements. In terms of the detection statistics (Table 3), the sensitivity of both methods improves with the number of time samples and the single tuning parameter method (3.24) results in higher power across all time samples. The method of regularizing with tuning parameters unique to each neighborhood (3.23) seems to perform similarly to the method when using a single penalty. The decomposition of the global detection statistics on a per vertex degree basis for the small-world network was also explored. Figure 7 a.), b.), and c.) represent the sensitivity, specificity, and probability of error, respectively, in reconstructing the neighborhoods of nodes as a function of node degree. The more highly connected nodes tend to have poorer sensitivity and higher probability of error. Figure 7 suggests that both methods tend to produce similar results in detection performance as a function of vertex degree. Given this similarity, one should opt for the reduced complexity of using single penalty with tuning parameter selected by (3.24).

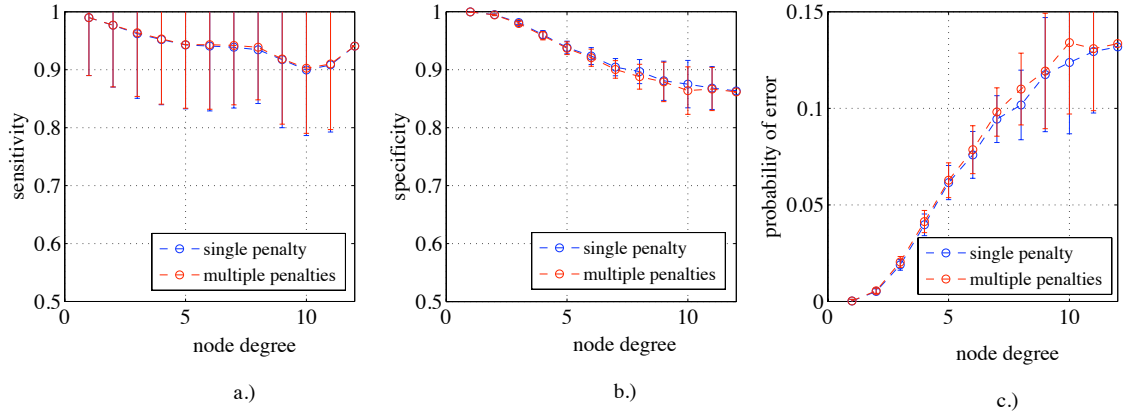

FIG 4. *Neighborhood detection statistics vs. node degree for 200 node scale-free network with $T = 1000$ and 1000 random trials. Initial condition was 40 randomly selected nodes as "infected" with rest "susceptible". a.) sensitivity, b.) specificity, c.) probability of error (red Single Penalty, blue Multiple Penalties)*

**5. Conclusion.** We have presented an estimator of the topology of interactions in a spatio-temporal graphical model. While the penalized likelihood formulation was derived for the general $SIR$ model, more complex $SIR$ processes, *i.e.*, $SI_1 \cdots, I_m RS$ could be handled by our approach. The detection performance resulting from simulations of a H1N1 epidemic model suggests that the proposed method accurately reconstructs the topology of

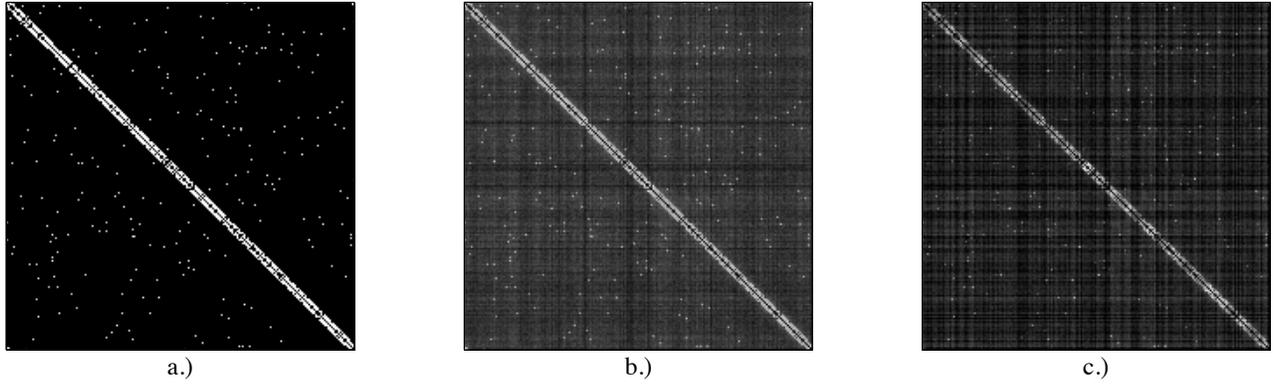

FIG 5.  *% zeros in the reconstruction of edges 200 node synthetic small world network under 100 time points resampled over 1000 initial conditions of 40 randomly selected nodes as "infected" with rest "susceptible". a.) ground truth, b.) single tuning parameter, c.) multiple tuning parameters (white - 0% black 100%)*

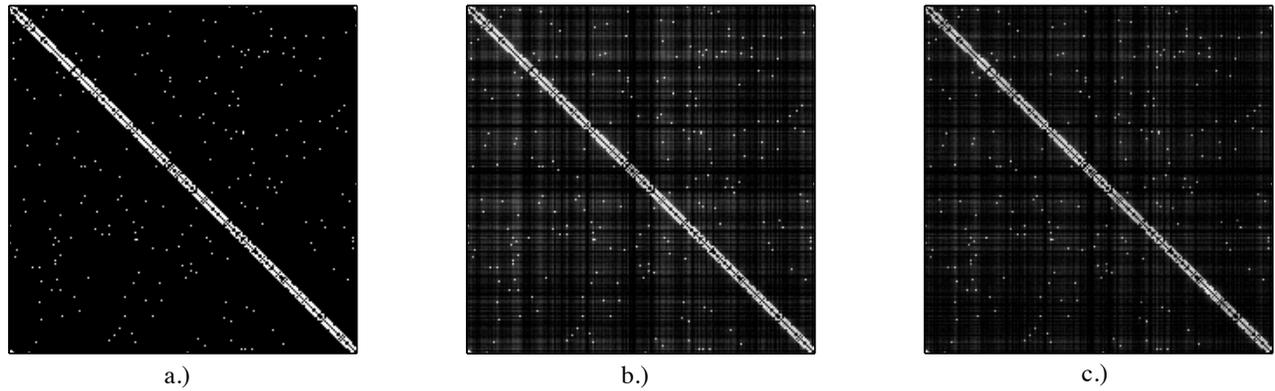

FIG 6.  *% zeros in the reconstruction of edges 200 node synthetic small world network under 1000 time points resampled over 1000 initial conditions of 40 randomly selected nodes as "infected" with rest "susceptible". a.) ground truth, b.) single tuning parameter, c.) multiple tuning parameters (white - 0% black 100%)*

TABLE 3

*Detection statistics vs. time horizon for 200 node synthetic small-world network with trajectories resampled over 1000 initial conditions of 40 randomly selected nodes as "infected" with rest "susceptible"*

| Method | $T$ | $Sens.(\lambda^*)$ | $Spec.(\lambda^*)$ | $P_e(\lambda^*)$ |
|---|---|---|---|---|
| $\ell_1$-SIR$(\lambda^*)$ | 100 | 0.26(0.05) | 0.94(0.01) | 0.08(0.01) |
| $\ell_1$-SIR$(\{\lambda_i^*\}_{i=1}^p)$ | 100 | 0.28(0.04) | 0.92(0.01) | 0.10(0.01) |
| $\ell_1$-SIR$(\lambda^*)$ | 400 | 0.41(0.02) | 0.95(0.00) | 0.07(0.00) |
| $\ell_1$-SIR$(\{\lambda_i^*\}_{i=1}^p)$ | 400 | 0.46(0.02) | 0.93(0.00) | 0.08(0.00) |
| $\ell_1$-SIR$(\lambda^*)$ | 700 | 0.77(0.02) | 0.90(0.01) | 0.11(0.01) |
| $\ell_1$-SIR$(\{\lambda_i^*\}_{i=1}^p)$ | 700 | 0.77(0.02) | 0.90(0.00) | 0.11(0.00) |
| $\ell_1$-SIR$(\lambda^*)$ | 1000 | 0.87(0.01) | 0.90(0.00) | 0.07(0.01) |
| $\ell_1$-SIR$(\{\lambda_i^*\}_{i=1}^p)$ | 1000 | 0.87(0.02) | 0.90(0.00) | 0.07(0.00) |

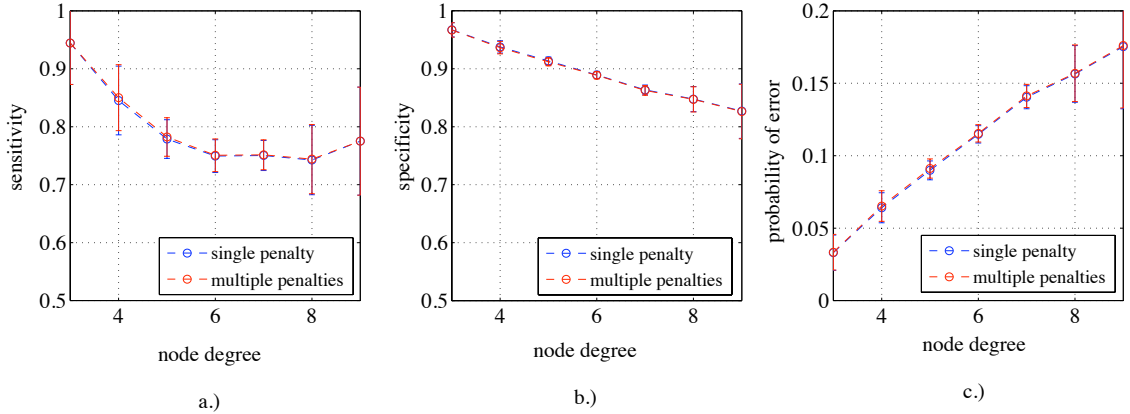

FIG 7. *Neighborhood detection statistics vs. node degree for 200 node small-world network with $T = 1000$ with trajectories resampled over 1000 initial conditions of 40 randomly selected nodes as "infected" with rest "susceptible". a.) sensitivity, b.) specificity, c.) probability of error (***red*** Single Penalty, ***blue*** Multiple Penalties)*

these types of networks while outperforming other state of the art structure learning algorithms.

**Acknowledgements.** The authors gratefully appreciate the insightful comments and discussions with Ami Wiesel, Kerby Shedden, and Ji Zhu.

## References.

Patrick L. Harrington Jr.
Bioinformatics Graduate Program
Department of Statistics
University of Michigan
2017 Palmer Commons Bldg.
100 Washtenaw Ave.
Ann Arbor, MI, 48109-2218
Tel. (734) 615-8895
FAX: (734) 615-6553
WWW: *http ://www.umich.edu/ ∼ plhjr*
E-mail: plhjr@umich.edu

Alfred O. Hero III.
Department of EECS
Department of Statistics
Department of Biomedical Engineering
University of Michigan
1301 Beal Avenue
Ann Arbor,MI 48109-2122
Tel. (734) 763-0564
FAX: (734) 763-8041
WWW: *http ://www.eecs.umich.edu/ ∼ hero*
E-mail: hero@umich.edu